\documentclass[letterpaper, 10 pt, conference]{ieeeconf}  

\IEEEoverridecommandlockouts                              

\usepackage{booktabs}
\usepackage{multirow}
\usepackage{cite}
\usepackage{todonotes}
\usepackage{comment}
\usepackage{amsmath,amssymb}
\usepackage{mathtools}
\usepackage{subcaption}

\title{\LARGE \bf
IntNet: A Communication-Driven Multi-Agent Reinforcement Learning Framework for Cooperative Autonomous Driving
}

\author{Leandro Parada$^{1}$, Kevin Yu$^{1}$, Panagiotis Angeloudis$^{1}$ 
\thanks{$^{1}$L. Parada, K. Yu, and P. Angeloudis are with the Centre for Transport Studies, Department of Civil and Environmental Engineering, Imperial College London, UK
        {\tt\small lap20@imperial.ac.uk}}%
}
\begin{document}

\maketitle
\thispagestyle{empty}
\pagestyle{empty}

\begin{abstract}

Achieving safety in autonomous driving through Multi-Agent Reinforcement Learning (MARL) is a critical yet challenging task due to non-stationarity, partial observability, and the need for effective coordination among agents. Although earlier cooperative MARL methods have aimed to improve coordination by sharing encoded state observations, we find these strategies are insufficient in safety-critical scenarios. To address this gap, we present IntNet, a novel MARL framework that enhances safety by incorporating the transmission of vehicle intent and adaptive communication scheduling into a unified end-to-end learning paradigm, jointly optimising all components for safe coordination.  Central to our approach is an observation prediction module that enables agents to forecast subsequent policy outputs, which they can then share across a vehicle-to-vehicle network. Our model-free architecture employs a Graph Attention Network to process incoming messages and a scheduler component to dynamically optimise the communication graph for bandwidth efficiency. We compare IntNet against state-of-the-art communicative MARL methods in complex urban environments with both autonomous and human-operated vehicles, achieving improved coordination, lower collision rates, and reducing communication efforts by up to 60\%. Through extensive experiments, we assess the framework's learning efficiency, scalability, and the balance between information sharing and bandwidth usage for collision-free trajectories.
\end{abstract}

\section{INTRODUCTION}

Deploying Connected and Autonomous Vehicles (CAVs) in dynamic urban environments requires effective coordination to ensure safety. While Multi-Agent Reinforcement Learning (MARL) has shown promise in coordinating autonomous tasks such as multi-robot path planning and navigation \cite{orr_multi-agent_2023}, applying it to autonomous driving presents distinct challenges. The complex and dynamic nature of urban settings and partial observability complicate the achievement of high levels of safety and coordination. Furthermore, the unpredictability of human driver behaviour adds a layer of complexity to this challenge.

\begin{figure}[t!] 
  \centering
  \includegraphics[trim={8cm 0 6cm 3cm},clip,scale=0.55]{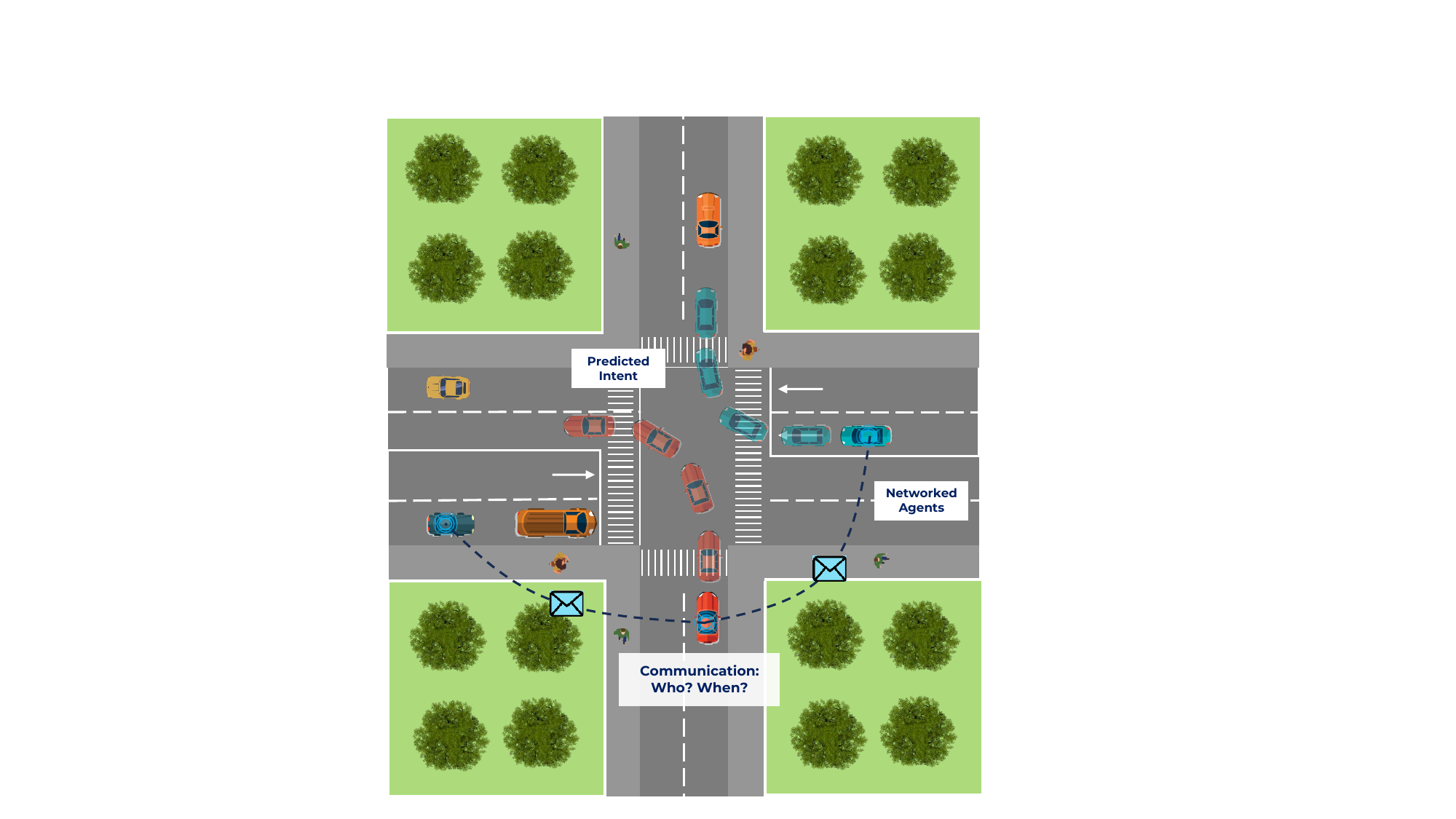}
  \caption{Illustration of our MARL framework in an urban traffic context, showcasing the strategic exchange of state information and vehicle intent, coordinated by a scheduler to optimise bandwidth and enable real-time decision-making for autonomous vehicles.}
  \label{fig:summary}
\end{figure}

Several applications of MARL in autonomous driving, spanning from intersection management to traffic optimisation, demonstrate its impact on the field. Notable research underscores MARL's efficacy in cooperative motion planning \cite{candela_transferring_2022, shalev-shwartz_safe_2016, troullinos_collaborative_2021}, as well as in controlling traffic signals to enhance urban mobility \cite{wang2021adaptive, wu2020multi, hu2023dynamic, wang2023traffic}. Despite these considerable advancements, achieving effective coordination and safety remains the most critical challenge for MARL in autonomous driving \cite{dinneweth_multi-agent_2022}.  While model-free MARL does not inherently provide safety guarantees, it offers a powerful methodology for addressing safety through trial-and-error learning, particularly in scenarios with diverse and complex configurations.

Agent coordination has been extensively studied within the MARL community by sharing state representations through a communication network \cite{zhu_survey_2022}. However, we argue that for safety-critical applications such as autonomous driving—-characterised by highly dynamic scenarios-—simply sharing state features is not enough. Understanding the future intentions of others provides additional contextual information to the agents that may otherwise be unavailable, thus unlocking opportunities for safer, more efficient decision making. Although some studies have explored intention-sharing in MARL \cite{jiang_learning_2018, chu_multi-agent_2020}, their applications are limited to simplistic grid-like environments, lacking the intricacies of autonomous driving, such as high-dimensional observations (e.g., Camera, LiDAR) and autonomous traffic dynamics.

To address the existing gap, we introduce IntNet, an intention-sharing MARL framework tailored for CAVs navigating complex urban environments safely and efficiently. IntNet operates through two communication steps. In the first step, agents share encoded messages and their recent decision-making strategies based on actual, observed information. In the second step, they communicate their estimated intentions over a future prediction horizon \(H\). Each agent uses an observation predictor to forecast environmental states, projecting future strategies over the horizon \(H\) to enhance coordination and proactive decision-making. To efficiently manage this process, a scheduler component is employed for bandwidth optimisation, while a graph attention network is used for processing incoming messages. The integration of each component within the end-to-end framework allows safety to emerge naturally from coordinated learning, rather than relying on hard constraints. The overall architecture employs an actor-critic loss alongside an independent supervised loss for training the observation prediction module. Training occurs directly from environmental interaction, eliminating the need for pre-training for the observation predictor. We benchmark IntNet against state-of-the-art Communicative MARL approaches across diverse urban scenarios and show how it can efficiently learn safer policies with low communication graph density.

The key contributions of this work are:
\begin{itemize}
    \item We propose a novel intention-sharing framework for communicative MARL, integrating Graph Attention Networks (GATs) for processing shared messages and a scheduling mechanism to optimise communication, enhancing coordination and reducing uncertainty in safety-critical scenarios.
    \item We develop a prediction module for multi-agent settings that forecasts future observations without requiring pretraining, allowing agents to recursively plan future actions.
    \item We advance beyond prior work in grid-based environments by incorporating high-dimensional observations and realistic vehicle kinematics, providing a more representative evaluation of communicative MARL in autonomous driving scenarios.
\end{itemize}


The paper is organised as follows: Section II offers a brief literature review on MARL for autonomous driving and recent advancements in Multi-Agent Communicative Reinforcement Learning. Section III details the paper's methodology, encompassing the cooperative MARL framework, training function, and environment description. Section IV discusses the experiments and key findings. Finally, Section V concludes the paper.

\section{RELATED WORK}

\subsection{MARL for Autonomous Driving}

The use of MARL has significantly impacted the field of autonomous driving, with its applications spanning from intersection management to traffic flow optimisation. A few survey works have helped categorise the literature \cite{dinneweth_multi-agent_2022, yadav_comprehensive_2023}. Recent efforts have demonstrated MARL's efficacy in facilitating collaborative navigation and path planning  \cite{candela_transferring_2022, shalev-shwartz_safe_2016, troullinos_collaborative_2021} and traffic signal control \cite{wang2021adaptive, hu2023dynamic, wang2023traffic}. These developments underline MARL's capacity to improve traffic efficiency through intelligent agent collaboration.

Further advancements in MARL for autonomous traffic management include addressing complex scenarios such as the double-merging problem \cite{shalev-shwartz_safe_2016}, cooperative lane-changing in the presence of Human-Driven Vehicles (HDVs) \cite{zhou_multi-agent_2022}, and highway ramp-merging strategies \cite{chen_deep_2022}. Key research areas focus on using Graph Convolutional Neural Networks (GCN) to enhance manoeuvring and safety \cite{dong_drl-based_2020} and applying MARL in challenging settings like near Emergency Vehicles \cite{parada_safe_2023} and in occluded situations \cite{parada_end--end_2023}. Despite these promising efforts, 
safety and coordination in dynamic urban environments persist as largely unresolved issues in MARL for autonomous driving, underscoring the need for innovative solutions to address the unpredictability and complexity of these scenarios \cite{dinneweth_multi-agent_2022}.

\subsection{Multi-Agent Communicative Reinforcement Learning}

Recent advances in Communicative Multi-Agent Reinforcement Learning (MACRL) highlight the importance of developing efficient communication methods among agents to enhance collaboration and save bandwidth. Early work by Foerster et al. \cite{foerster_learning_2016} and Sukhbaatar et al. \cite{sukhbaatar_learning_2016} introduced the concept of MARL agents learning to communicate for group decision-making, which was later refined by Jiang et al. \cite{jiang_learning_2018} and Singh et al. \cite{singh_learning_2018}, who introduced gating mechanisms for selective communication based on immediate surroundings. Further innovations include TarMAC by Das et al. \cite{das_tarmac_2020}, which uses attention mechanisms for focused communication, and work by Wang et al. \cite{wang_learning_2020} and Liu et al. \cite{liu_multi-agent_2019} on communication graphs to manage bandwidth and improve precision. Niu et al.'s MAGIC system \cite{niu_multi-agent_2021} represents a significant advancement, dynamically adjusting communication based on agents' conditions and environmental factors.

Intention-sharing in MARL has also been explored, with studies like those by Jiang et al. \cite{jiang_learning_2018} and Chu et al. \cite{chu_multi-agent_2020} demonstrating the benefits of agents sharing actions or policy outputs to reduce system non-stationarity. Kim et al. \cite{kim_communication_2020} further improved cooperation through intention-sharing using observation and action prediction modules with attention mechanisms, achieving superior performance in grid-based environments. Wang et al. \cite{wang_too_many} introduced a Bayesian method for decentralised multi-agent collaboration, enabling agents to infer hidden intentions through inverse planning and coordinate both high-level plans and low-level actions. Building on these approaches, our research introduces an intention-sharing MARL framework tailored for autonomous driving, employing Graph Attention Networks (GAT) and a message scheduling mechanism to handle the complexities of high-dimensional observations and communication bandwidth constraints effectively.

\section{METHODOLOGY}

\begin{figure*}[t]
  \centering
  \includegraphics[width=0.7\textwidth, trim={0 0cm 0 3cm}]{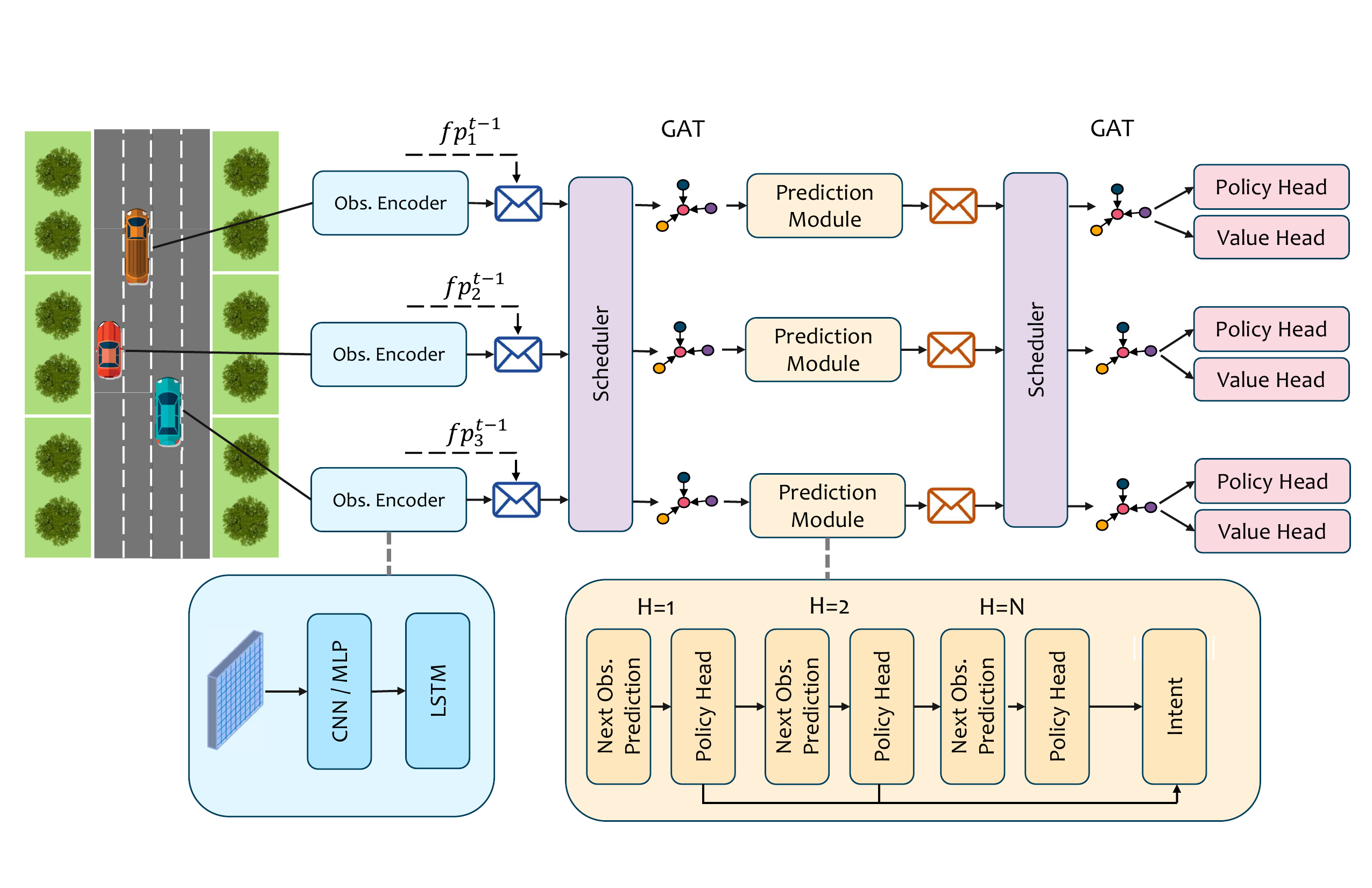}
  \caption{This figure depicts IntNet, our MARL framework for safe, efficient CAV navigation, highlighting the sharing of encoded observations and vehicle intentions. The scheduler optimises the communication graph, while the GAT module processes incoming messages with attention to prioritise information for decision-making.}
  \label{fig:framework}
\end{figure*}

\subsection{Communication-based Multi-Agent Reinforcement Learning}

We formulate the MACRL as a Decentralised Partially Observable Markov Decision Process (Dec-POMDP) \cite{Kaelbling1998PlanningAA}, represented by the tuple \( \langle N, S, A, O, \mathcal{T}, R, \gamma, \mathcal{M}, \mathcal{C} \rangle \). Here, \( N = \{1, 2, \ldots, n\} \) is the number of agents. \( S \) denotes all possible states, where \( s^t \in S \) represents the environment's state at time \( t \). The set \( A \) comprises all possible actions, with each agent \( i \) selecting an action \( a_i^t \in A \) at each time step.

Agents receive partial information via observations \( o_i^t \in O \), and the transition between states, given agents' actions, is defined by the transition function \( \mathcal{T}: S \times A \rightarrow S \). The reward function \( R: S \times A \times S \rightarrow \mathbb{R} \) determines the immediate reward. Current and future rewards are balanced through the discount factor \( \gamma \in [0, 1] \).

Communication occurs through messages \( \mathcal{M} \) and is governed by the protocol \( \mathcal{C}: \mathcal{M}^N \rightarrow \mathcal{M} \), which defines how messages are aggregated. The overall objective is to determine policies \( \pi_i: O \times \mathcal{M} \rightarrow A \) for each agent that maximise the expected cumulative reward: \( \max_{\pi_1, \pi_2, \ldots, \pi_N} \mathbb{E} \left[ \sum_{t=0}^{T} \gamma^t r^t(s^t, a_{1}^t, a_{2}^t, \ldots, a_{N}^t, s^{t+1}) \right] \), where \( r^t(s^t, a^t, s^{t+1}) \) is the immediate reward at time \( t \), defined by \( R(s^t, a^t, s^{t+1}) \).

\subsection{Intention-sharing MARL framework}

Our proposed Intention-sharing MARL framework, as illustrated in Figure~\ref{fig:framework}, is designed to facilitate communication among agents in a cooperative setting using a communicative actor-critic model. This framework operates through two communication steps: This framework operates through two communication steps: one for sharing current observations and policy fingerprints, and another for exchanging estimated intentions over a future prediction horizon.

\subsubsection{Observation Encoding and Policy Fingerprint} 

In the first step, each agent \(i\) encodes its local observation \(o_i^t\) at time step \(t\) into a message \(m_i^t\) using an encoding function \(f_{\text{enc}}: O \rightarrow \mathcal{M}\). The encoded message \(m_i^t = f_{\text{enc}}(o_i^t)\) effectively summarises the observed state information that can be shared with other agents. Alongside this, each agent also broadcasts its policy fingerprint from the previous time step, denoted by \(\mathbf{fp}_i^{t-1}\). This fingerprint summarises the agent’s recent decision-making context and serves as a signature of its strategy. The policy fingerprint is defined as the probability distribution over the action space \(A\):

\begin{equation}
\mathbf{fp}_i^{t-1} = \left\{ \pi_{\theta}(a \mid o_i^{t-1}, m_{-i}^{t-1}) \mid a \in A \right\},
\end{equation}

Where \(o_i^{t-1}\) represents the local observation of agent \(i\) at time step \(t-1\), and \(m_{-i}^{t-1}\) represents the messages received from all other agents at time step \(t-1\). 
This approach enables agents to share both their current observations and their recent decision-making context, establishing a foundation for collaborative action planning.

\subsubsection{Intention Prediction and Sharing} In the second step, each agent predicts and shares its future intentions over a specified time horizon \(H\). These predictions are represented by a sequence of future policy fingerprints \(\mathbf{I}_i^{t+H}\), which summarise the agent's projected decision-making strategies for the upcoming \(H\) steps. The intention set \(\mathbf{I}_i^{t+H}\) is formulated as:

\begin{equation}
    \mathbf{I}_i^{t+H} = 
    \begin{bmatrix}
        \mathbf{fp}_i^{t+H}\\
        \mathbf{fp}_i^{t+H-1} \\
         \vdots \\
        \mathbf{fp}_i^{t+1}
    \end{bmatrix}
\end{equation}

where each future policy fingerprint \(\mathbf{fp}_i^{t+h}\) for \(h = 1, 2, \ldots, H\) is computed as:

\begin{equation}
\mathbf{fp}_i^{t+h} = \left\{ \pi_{\theta}(a \mid \hat{o}_i^{t+h}, m_{-i}^{t+h}) \mid a \in A \right\}.
\end{equation}

In this equation, \(\hat{o}_i^{t+h}\) represents the predicted observation of agent \(i\) at time step \(t+h\), which is derived using an observation prediction module \(f_{\text{obs}}\). The predicted observation \(\hat{o}_i^{t+h}\) is calculated recursively starting from the agent’s current observation \(o_i^t\) and the policy fingerprints from the other agents. The recursive prediction is defined as:

\begin{equation}
    \hat{o}_i^{t+h} = \begin{cases}
        f_{\text{obs}} \left( o_i^{t},\mathbf{fp}_{-i}^{t-1} \right) & \text{if} \quad h = 1\\
        f_{\text{obs}} \left( \hat{o}_i^{t+h-1}, \mathbf{fp}_{-i}^{t+h-1} \right) & \text{if} \quad h>1\\
    \end{cases}
\end{equation}

where \(\mathbf{fp}_{-i}^{t+h-1}\) denotes the collective policy fingerprints from all agents except \(i\) at time step \(t+h-1\). At the first step (\(h = 1\)), the function \(f_{\text{obs}}\) uses the agent’s current observation \(o_i^t\) and the policy fingerprints from the other agents at the previous time step \(\mathbf{fp}_{-i}^{t-1}\). This initial prediction sets the stage for subsequent recursive steps.

For \(h > 1\), \(f_{\text{obs}}\) recursively updates the predicted observation by taking as input the predicted observation from the previous step \(\hat{o}_i^{t+h-1}\) along with the corresponding policy fingerprints \(\mathbf{fp}_{-i}^{t+h-1}\). Through this recursive mechanism, \(f_{\text{obs}}\) effectively captures the evolving dynamics of the environment, informed by the continuous inte between the agent's own evolving state and the intentions of other agents. This allows each agent to form a coherent and forward-looking plan of its anticipated observations over the horizon \(H\), thus enabling more informed and coordinated decision-making.

\subsubsection{Communication Scheduling and Graph Attention} Given limited bandwidth and the high-dimensional state space in autonomous driving, our approach includes a scheduler for each communication step. This scheduler determines the communication links among agents at each time step \( t \) by constructing an adjacency matrix \( G^t \). The element \( G_{ij}^t \) of this matrix is 1 if agent \( i \) communicates with agent \( j \) at time \( t \), and 0 otherwise. This setup allows for optimising communication by reducing unnecessary data exchange, thereby enhancing bandwidth efficiency.




Agents use a Graph Attention Network (GAT) following each communication step to merge and process the received messages.  The GAT employs an attention mechanism to dynamically weigh the importance of each message, allowing agents to prioritise the most relevant information. This prioritization is crucial for effective decision-making, particularly in high-dimensional, bandwidth-constrained environments where not all information can be equally considered. This process is mathematically represented as follows. Given a node $i$ with feature representation $h_i$ and its neighbors $\mathcal{N}(i)$, the attention coefficient $\alpha_{ij}$ between node $i$ and its neighbor $j$ is computed as:

\begin{equation}
\alpha_{ij} = \frac{\exp\left( \text{LeakyReLU}\left( \mathbf{a}^T [\mathbf{W}h_i \| \mathbf{W}h_j] \right) \right)}{\sum_{k \in N(i) \cup \{ i \}} \exp\left( \text{LeakyReLU}\left( \mathbf{a}^T [\mathbf{W}h_i \| \mathbf{W}h_k] \right) \right)}
\end{equation}

where $\|$ denotes concatenation, $\mathbf{W}$ is a weight matrix applied to each node, $\mathbf{a}$ is a weight vector for the attention mechanism, and LeakyReLU is a nonlinear activation function. The attention coefficients $\alpha_{ij}$ are normalised across all neighbours $k$ of $i$ using the softmax function, ensuring comparability.

The updated feature representation \( h'_i \) of node \( i \), incorporating information from its neighbours, is then calculated as a weighted sum: \( h'_i = \sigma\left( \sum_{j \in \mathcal{N}(i) \cup \{ i \}} \alpha_{ij} \mathbf{W}h_j \right) \), where \( \sigma \) denotes an optional activation function. This formulation allows the GAT to adaptively prioritise information from the most relevant neighbours, efficiently integrating transmitted messages into meaningful features for enhanced decision-making. Finally, once the agents have processed the incoming messages, they determine the policy and value outputs through independent linear layers as depicted in figure \ref{fig:framework}.

\subsection{Training}

observation prediction module \(f_{\text{obs}}\) and the policy are trained concurrently using data collected from episode rollouts. In contrast to previous approaches such as Nagabandi et al. \cite{nagabandi_neural_2017}, which pre-train the predictor, we do not pre-train \(f_{\text{obs}}\). Instead, transitions stored during the rollouts are used to update both the prediction module and the policy simultaneously.

To train the observation prediction module, each episode step generates transitions \(\tau = (o^t, a^t, o^{t+1})\), from which we compile a supervised dataset with inputs \(o^t, a^t\) and targets \(o^{t+1}\). Observations are normalised, and variables are categorised as continuous or categorical. \(f_{\text{obs}}\) is trained using the Mean Squared Error (MSE) loss:

\begin{equation} 
\mathcal{L}_{obs} = \frac{1}{M}\sum_{i=1}^{M}(o_i^{t+1} - \hat{o}_i^{t+1})^2,
\end{equation}

where \(o_i^{t+1}\) represents the actual next observation, \(\hat{o}_i^{t+1}\) denotes the predicted next observation, and \(M\) is the total number of samples.

In parallel, the policy is trained using an actor-critic loss, which incorporates the losses of both the actor (policy) and the critic (value):

\begin{equation}
\begin{aligned}
\mathcal{L}_{AC} = & \mathbb{E}_{(\tau)} \left[ \sum_{t=0}^{T} A^t \cdot \nabla_{\theta} \log \pi_i (a_i^t | o_i^t, m_i^t)  \right] \\
& + \lambda \sum_{t=0}^{T} \nabla_{\rho}(R_{i}^{t}- V_{\rho}(o_{i}^{t}))^2,
\end{aligned}
\end{equation}

where \( A^t = R_i^t - V_{\rho}(o_i^t) \) is the advantage function, and \( V_{\rho}(o_i^t) \) is the value function.

This concurrent training setup allows the observation prediction module to improve the policy's understanding of environment dynamics in real time, without requiring a separate pre-training phase.

\subsection{Environment Description}

\begin{figure}[t]
\centering
\begin{subfigure}[t]{0.42\textwidth}
    \centering
    \includegraphics[trim={2.5cm 3cm 9cm 5cm},clip,width=\textwidth]{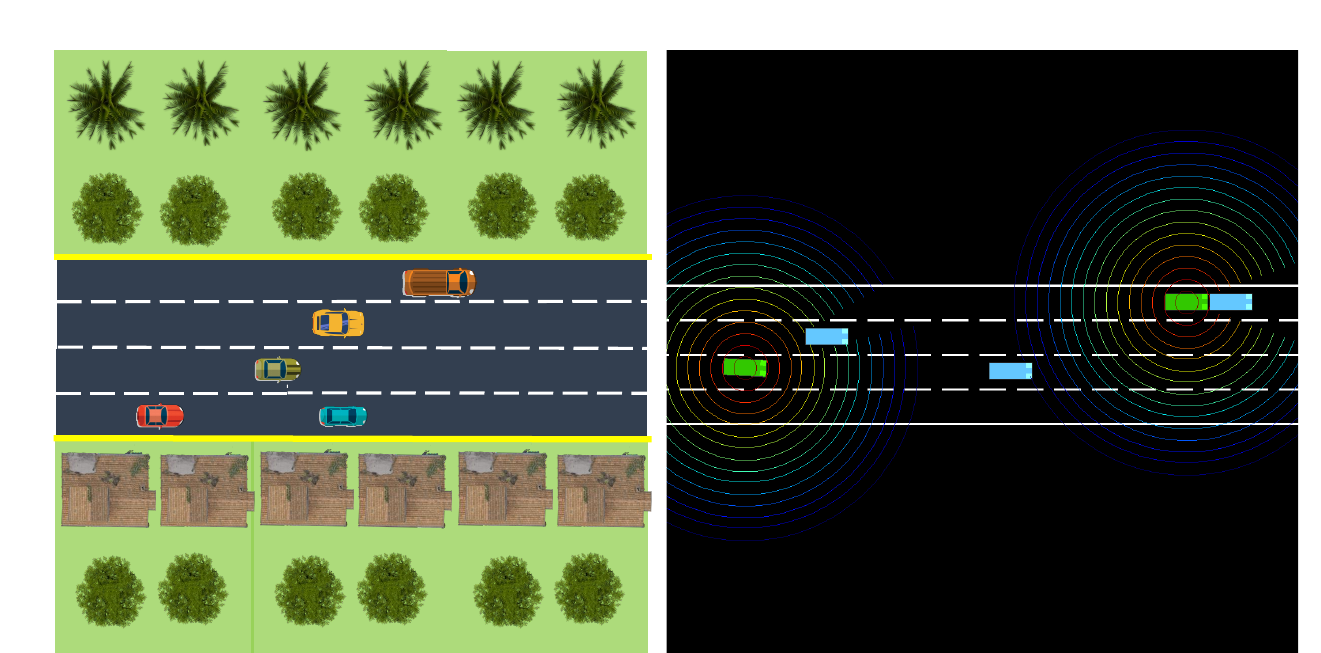}
    \caption{Highway.}
    \label{fig:highway}
\end{subfigure}
\hspace{10mm}
\begin{subfigure}[t]{0.42\textwidth}
    \centering
    \includegraphics[trim={4cm 5cm 11cm 6.0cm},clip, width=\textwidth]{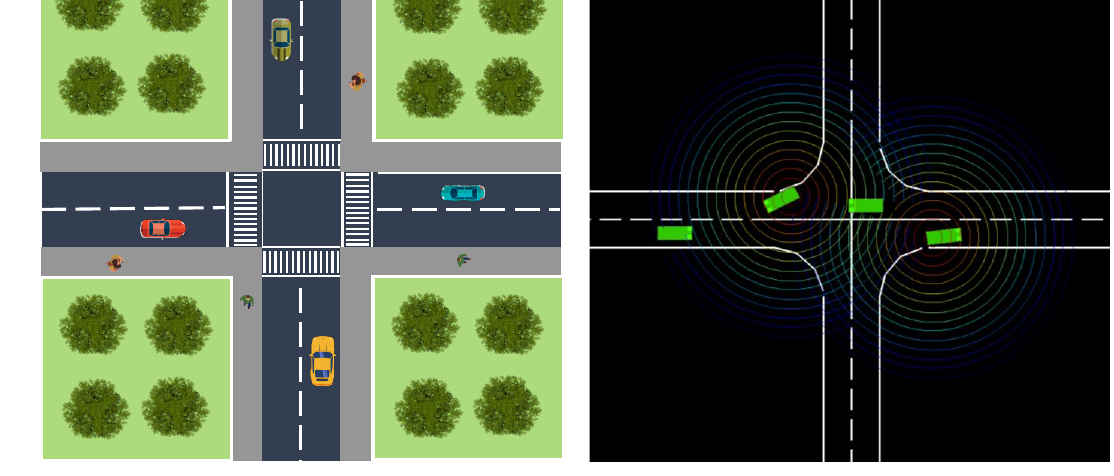}
    \caption{Intersection.}
    \label{fig:intersection}
\end{subfigure}
\hspace{10mm}
\begin{subfigure}[t]{0.47\textwidth}
    \centering
    \includegraphics[trim={-1.5cm 5cm 1cm 7cm},clip, width=\textwidth]{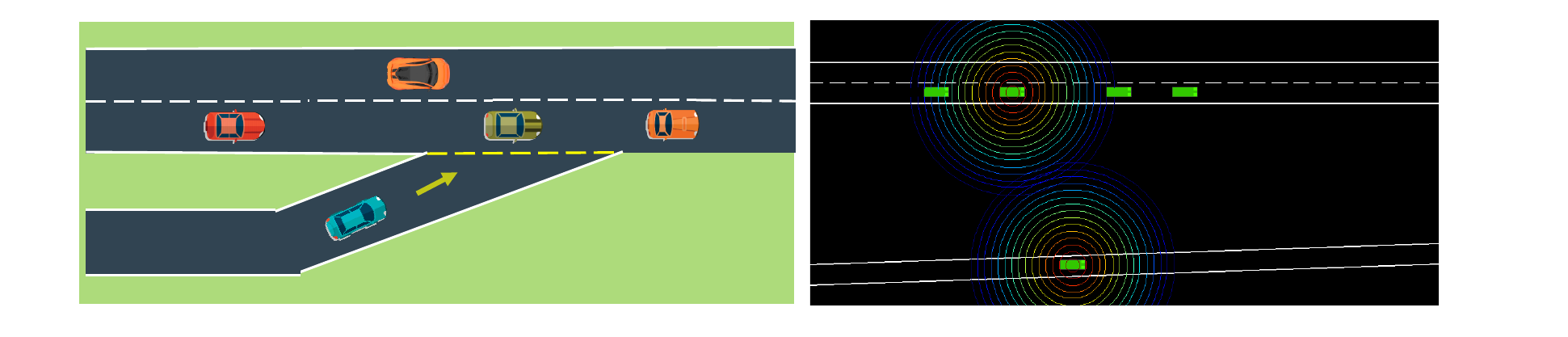}
    \caption{Lane merging.}
    \label{fig:merging}
\end{subfigure}
\caption{Scenarios used for the experiments.}
\label{fig:scenarios}
\end{figure}

\begin{figure*}
\centering
\begin{subfigure}{0.32\textwidth}
    \centering
    \includegraphics[width=\textwidth]{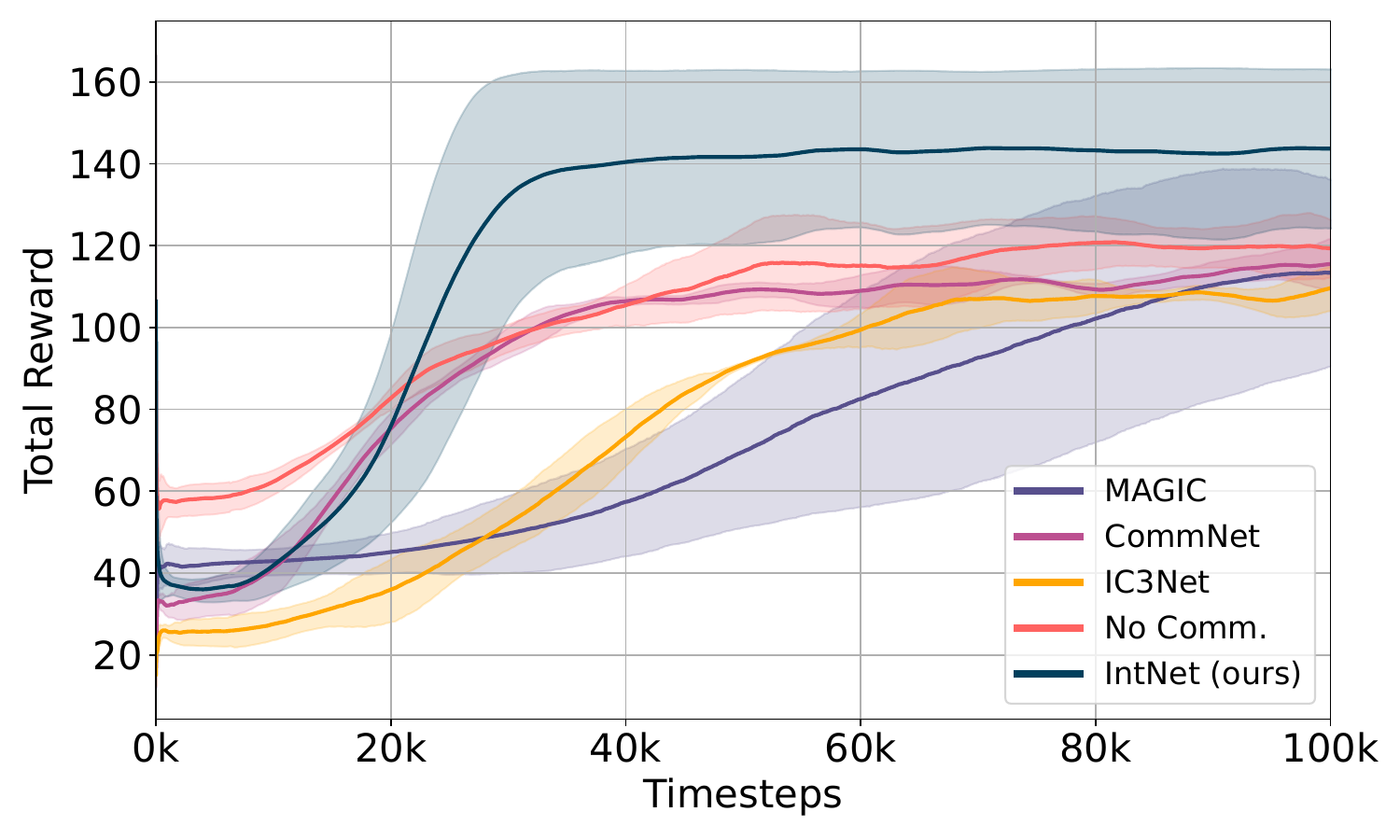}
    \caption{Highway.}
    \label{fig:TJ_5_agents}
\end{subfigure}
\begin{subfigure}{0.32\textwidth}
    \centering
    \includegraphics[width=\textwidth]{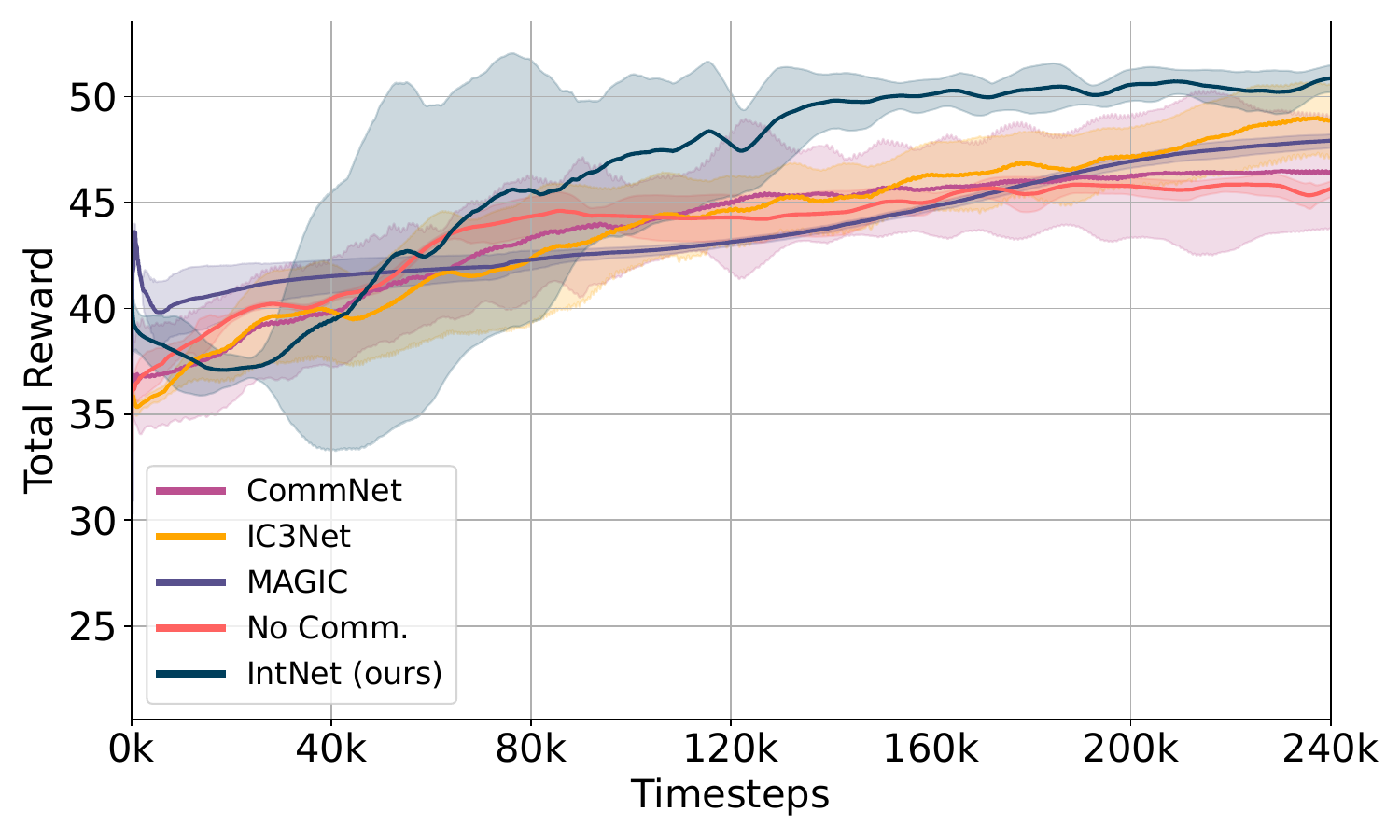}
    \caption{Intersection}
    \label{fig:another_subfigure}
\end{subfigure}
\begin{subfigure}{0.32\textwidth}
    \centering
    \includegraphics[width=\textwidth]{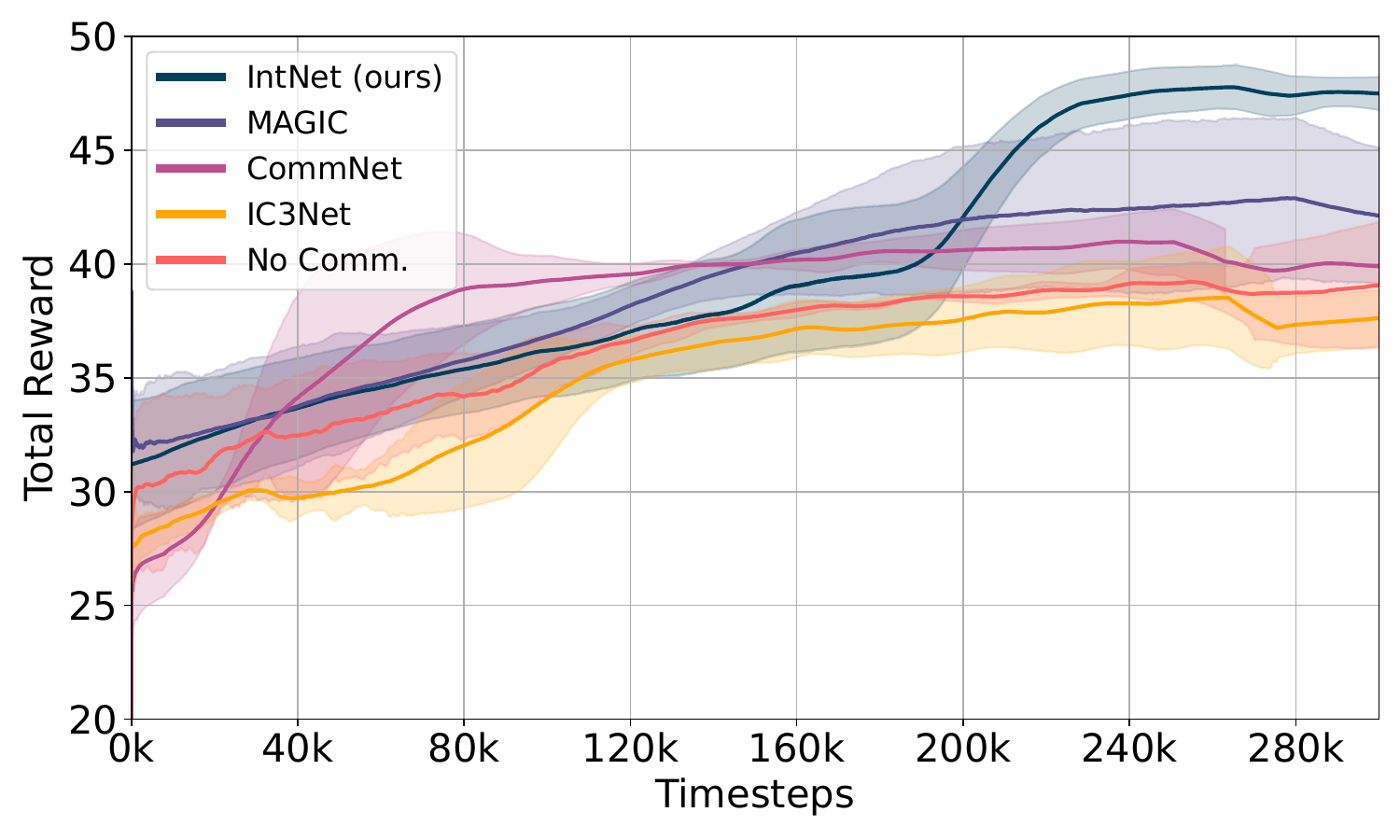}
    \caption{Lane Merging}
    \label{fig:another_subfigure}
\end{subfigure}
\caption{Team reward of the different methods during training.}
\label{fig:combined_figure_return}
\end{figure*}

\begin{figure*}
\centering
\begin{subfigure}{0.32\textwidth}
    \centering
    \includegraphics[width=\textwidth]{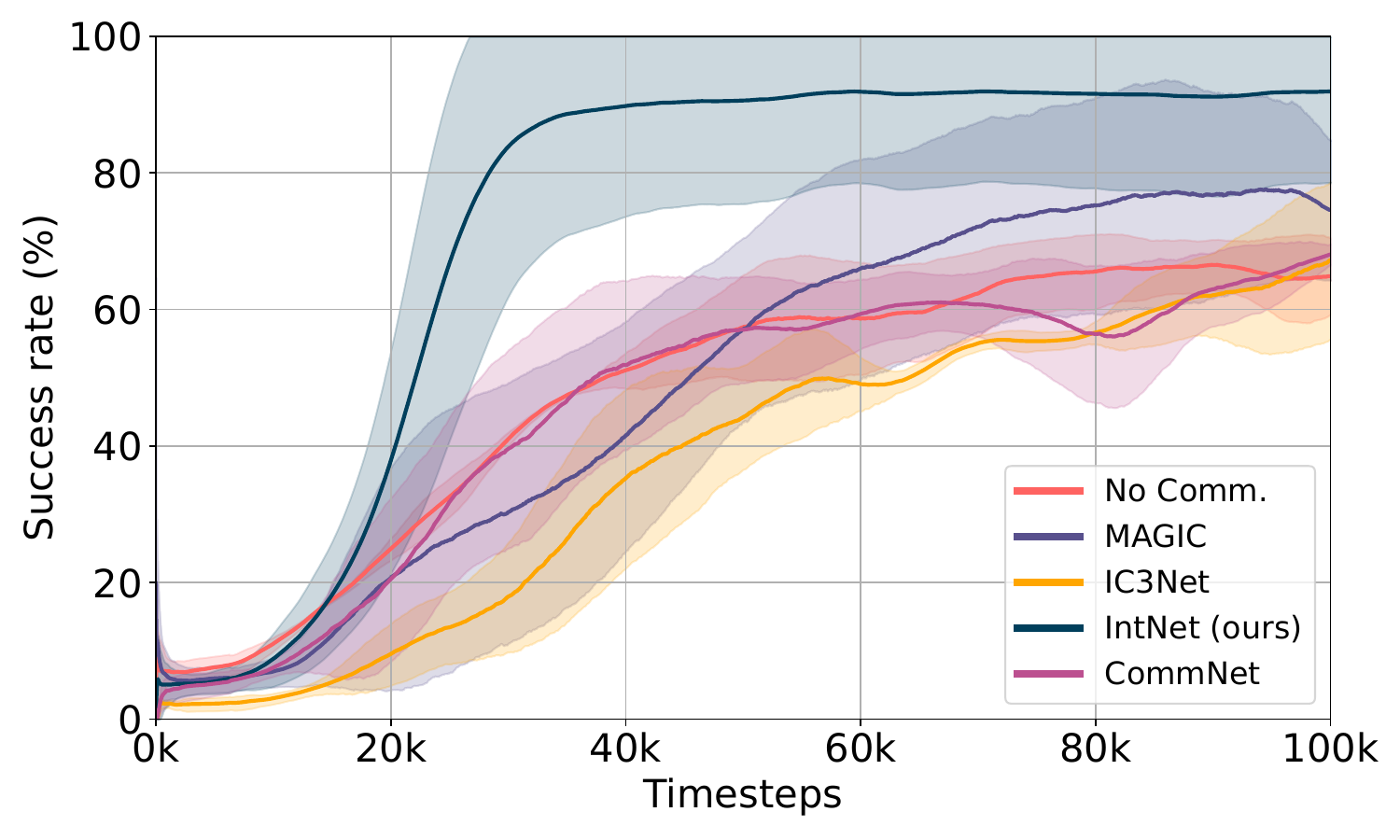}
    \caption{Highway.}
    \label{fig:TJ_5_agents}
\end{subfigure}
\begin{subfigure}{0.32\textwidth}
    \centering
    \includegraphics[width=\textwidth]{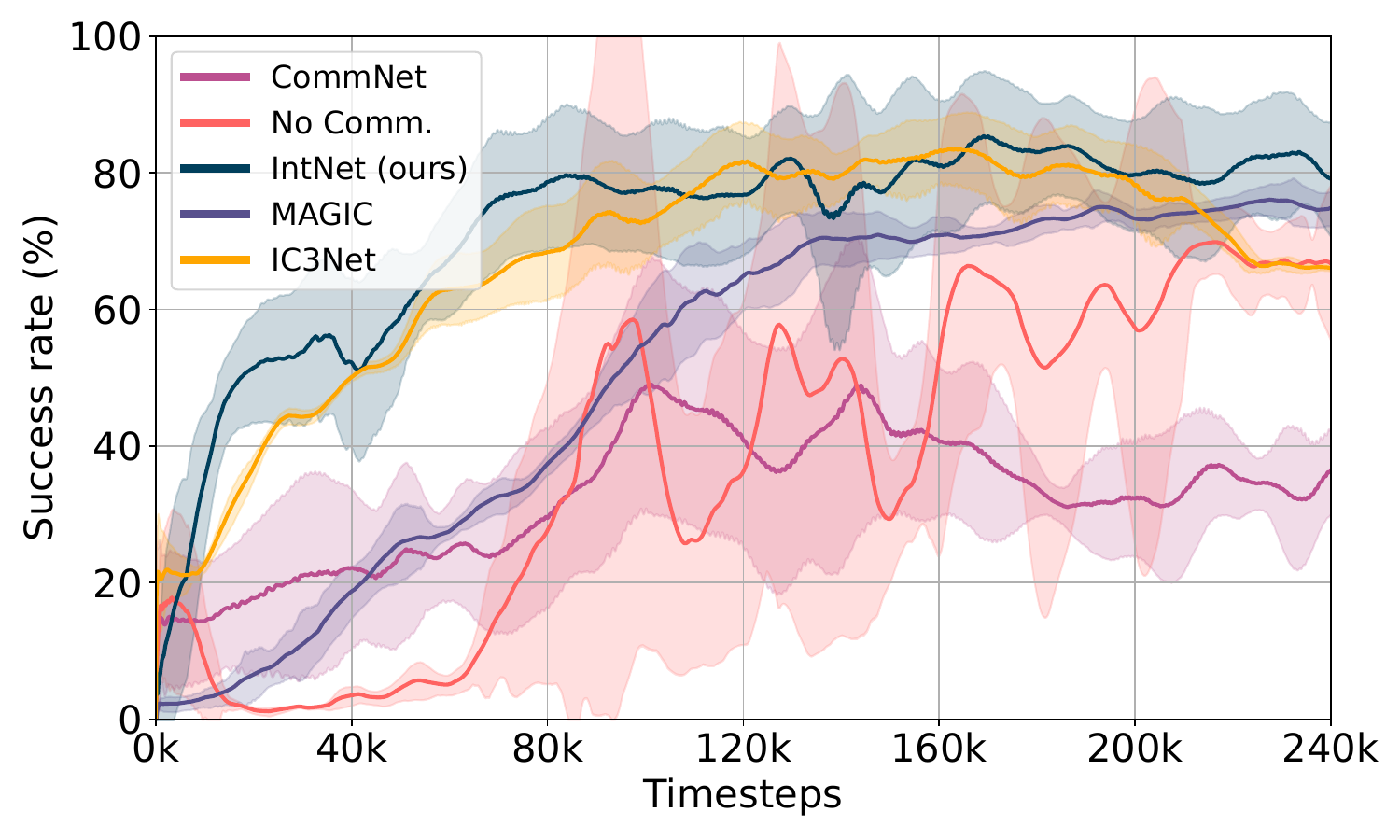}
    \caption{Intersection}
    \label{fig:another_subfigure}
\end{subfigure}
\begin{subfigure}{0.32\textwidth}
    \centering
    \includegraphics[width=\textwidth]{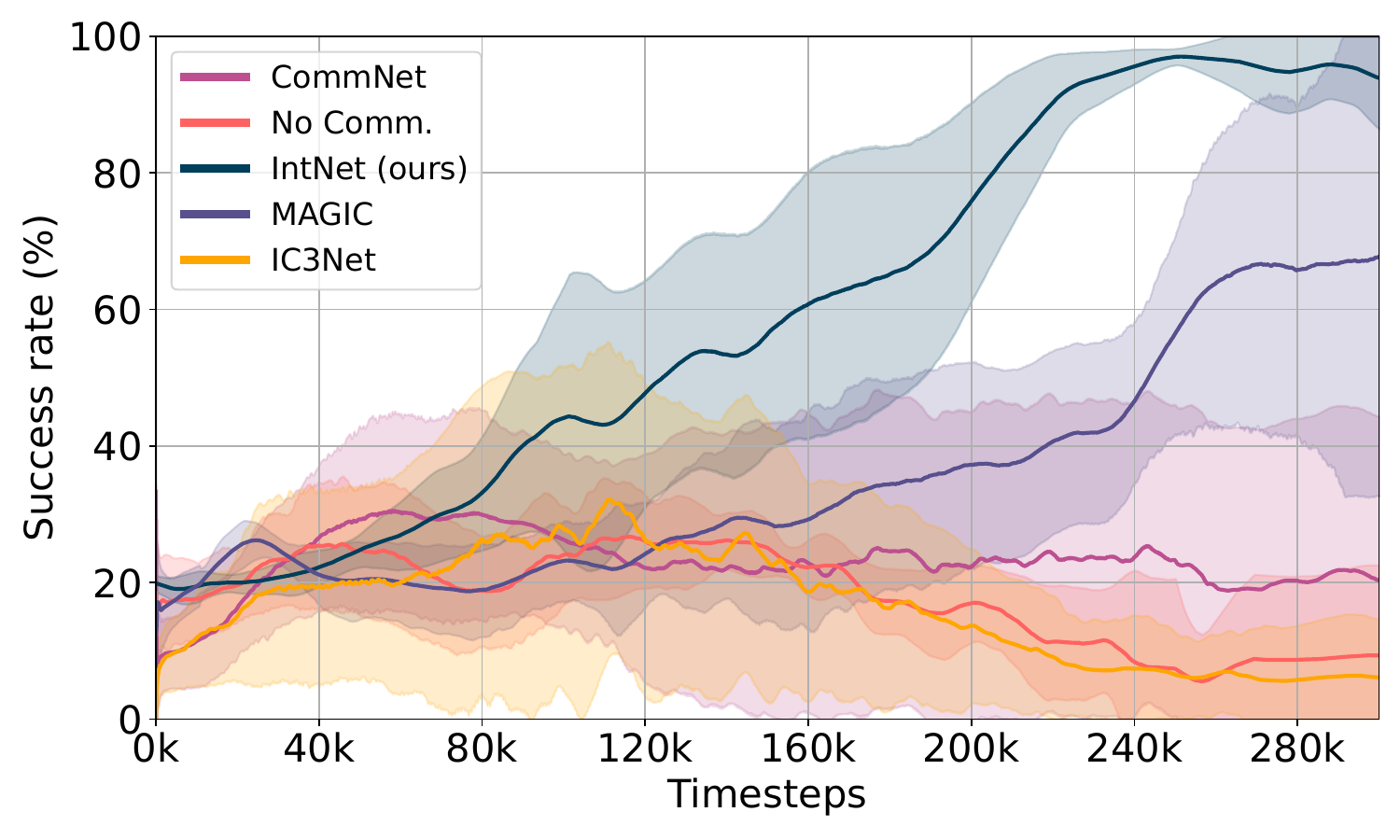}
    \caption{Lane Merging}
    \label{fig:another_subfigure}
\end{subfigure}
\caption{Success rate of the different methods during training.}
\label{fig:combined_figure_arrived}
\end{figure*}

 We use a modified version of Highway-env \cite{highway-env}, adapted for multi-agent urban navigation where agents avoid collisions and reach their destinations to maximise rewards. While Highway-Env serves as the foundational simulator, the addition of real-time communication capabilities and dynamic graph learning significantly increases the complexity of the scenarios. Agents must process communication in real-time, construct dynamic graphs for decision-making, and coordinate in a multi-agent setting, which presents substantial computational and learning challenges. To enhance adaptability, we introduce randomness in vehicles' speeds, locations, and destinations, and limit agent communication to 60 meters to align with DSRC protocol constraints.

The elements of the gym environment can be summarised as follows: 
\begin{itemize}
    \item Agents: CAVs with limited visibility, capable of exchanging messages through a vehicle-to-vehicle communication network.
    \item Observation: LiDAR-like method using raycasting to segment the 360° field around the agent, detecting other vehicles within an 80-meter range. We use a total of 64 rays and update the observation every timestep.
    \item Action: Discrete meta-actions including acceleration, deceleration, speed maintenance, and lane changes (left and right).  The speed values for each scenario are included in the Appendix.
\item Reward: The reward function is designed for a cooperative game and incorporates components for speed ($r_{\text{speed}}$), collision avoidance ($r_{\text{col}}$), arrival ($r_{\text{arr}}$), and bonuses for cooperative lane merging ($r_{\text{merge}}$). To balance these components, we performed a preliminary grid search over weight configurations, ensuring safety was prioritised above all else. Within configurations that met safety requirements, we tuned other components, such as $r_{\text{speed}}$, to encourage efficient driving. Additionally, we tailored the reward weights to specific scenarios, as certain components, such as $r_{\text{arr}}$, are less relevant in environments like highway driving. This approach ensures that the reward function aligns with the unique objectives of each scenario while promoting coordination and safety. The specific values for each are included in the Appendix.
\end{itemize}

\section{Experiments}

We benchmark our method against baselines across three scenarios: Highway, 4-way Intersection, and Lane Merging, depicted in Figure \ref{fig:scenarios}, which includes both schematic and actual environment representations to illustrate the observations. The Highway and Lane Merging environment contain both CAVs and human-driven vehicles (HDV), while the Intersection environment consists of only CAVs. To model the HDVs, we use the IDM/MOBIL car following model \cite{kesting_2007}, which has been extensively used in the literature to represent human-driven behaviour with a simple analytical model. 

The primary evaluation metrics in our study are the average arrival and collision rates, expressed in percentages. The arrival rate quantifies the percentage of episodes in which all CAVs successfully reached their destinations, while the collision rate measures the percentage of episodes involving at least one CAV collision. Additionally, the average team return is evaluated to assess overall performance. To ascertain the robustness of our training outcomes, we utilise three independent seeds to generate learning curves and validate our results over 100 test episodes. The prediction horizon for all experiments was set to $H=2$ based on a preliminary experiment, the results of which are provided in the Appendix.

\subsection{Baseline comparison}

We evaluated our method's performance against established communication-based MARL approaches, which primarily focus on sharing state or state features. The comparison included three baseline methods: No Communication, CommNet \cite{sukhbaatar_learning_2016}, IC3Net \cite{singh_learning_2018} and MAGIC \cite{niu_multi-agent_2021}.

Figures \ref{fig:combined_figure_return} and \ref{fig:combined_figure_arrived} illustrate the team reward and arrival statistics, respectively, across three scenarios for all methods, with IntNet consistently achieving higher rewards and success rates. Variations in arrival rates indicate that intention-sharing is particularly beneficial in specific scenarios, such as lane merging where it's crucial for safe navigation, while its impact in the intersection scenario is smaller. Table \ref{baseline_comparison} further details the arrival rate (\%) test performance over 100 independent episodes in these scenarios, supporting these observations.

In the highway scenario, our proposed method enables agents to reach high success levels with just 100k training episodes, compared to baselines, which require around 200k episodes for similar success and safety outcomes. In the intersection scenario, although IC3Net shows success comparable to our IntNet method, ours provides a marginally quicker and more consistent learning curve. We see that all baseline algorithms underperform in the lane-merging scenario. This scenario demands the most cooperation out of the three, causing baseline algorithms to learn inconsistently and often favour solutions that increase the speed at the expense of safety, leading to more collisions.



\begin{table}
\centering
\caption{Test arrival rate (\%) of different Communicative MARL methods.}
\label{baseline_comparison}
\begin{tabular}{lccc}
\hline
Method & Highway & Intersection & Lane Merging \\
\hline
No Comm & \(62.6 \pm 3.6 \) & \(65.3 \pm 8.4 \) & \(10.3 \pm 6.4 \) \\
IC3Net & \(59.5 \pm 6.1 \) & \(70.7 \pm 6.3 \) & \(10.5 \pm 8.3 \)\\
CommNet & \(64.1 \pm 8.4 \) & \(41.0 \pm 8.2 \) & \(26.3 \pm 7.6 \)\\
MAGIC & \(77.6 \pm 8.1\) & \(75.4 \pm 6.7 \) & \(66.1 \pm 10.1 \) \\
IntNet (Ours) & \(97.1 \pm 2.6\) & \(89.7 \pm 5.9 \) & \(93.2 \pm 9.9 \) \\
\hline
\end{tabular}
\end{table}

\subsection{Scalability}

\begin{figure}[]
\centering
\begin{subfigure}[]{0.2\textwidth}
    \centering
    \includegraphics[trim={0cm 0cm 0cm 10cm},clip,width=\textwidth]{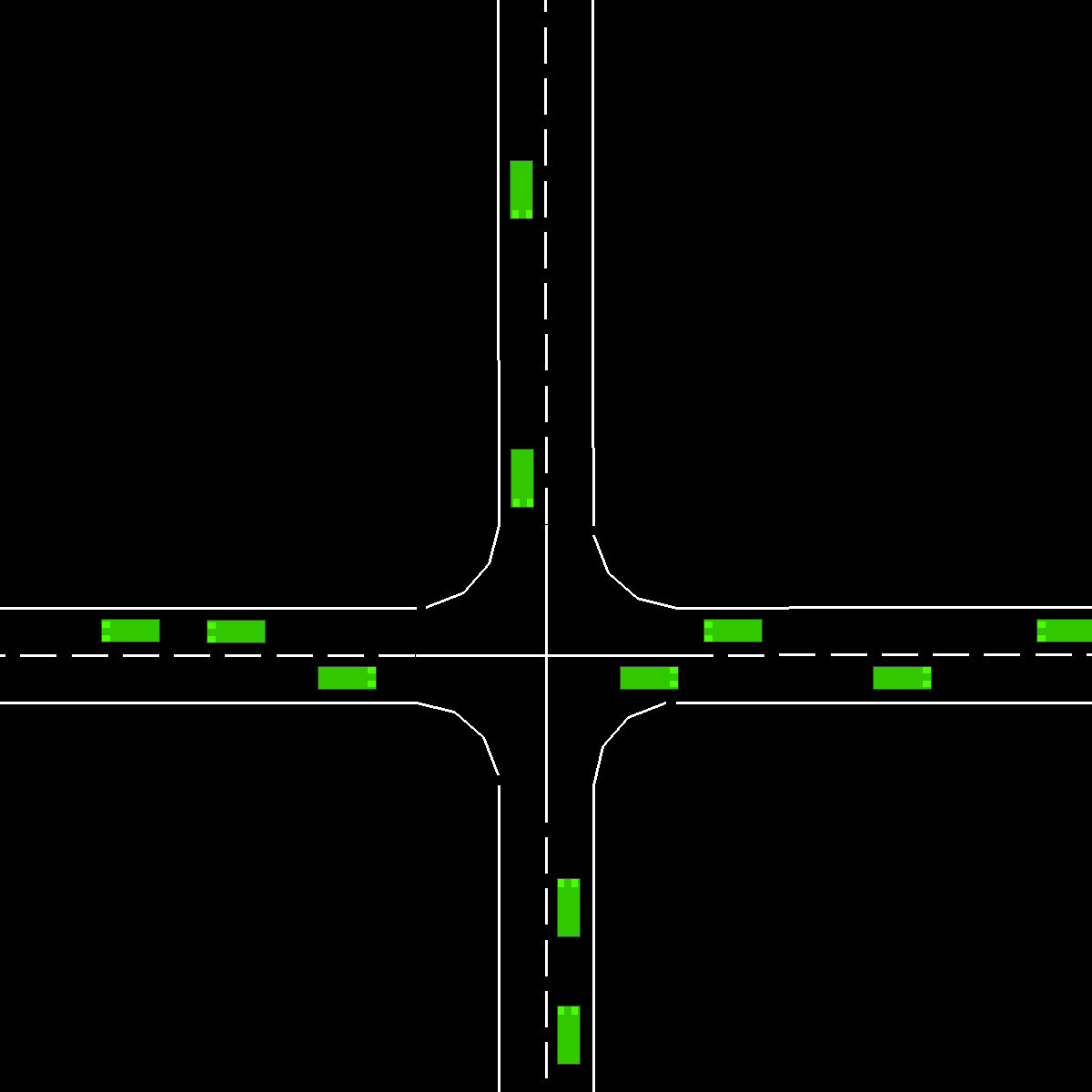}
    \caption{}
    \label{fig:highway}
\end{subfigure}
\begin{subfigure}[]{0.26\textwidth}
    \centering
    \includegraphics[width=\textwidth]{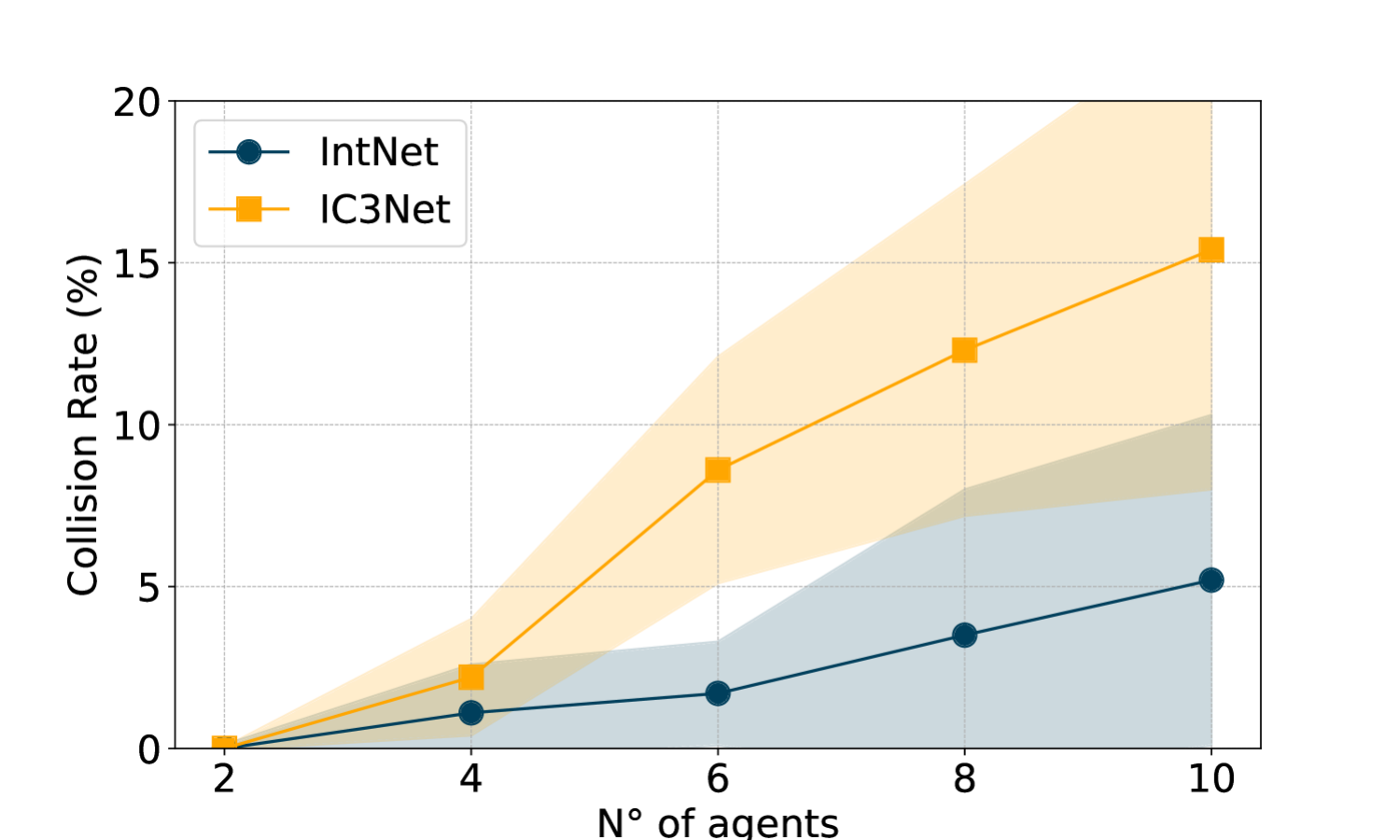}
    \caption{}
    \label{fig:intersection}
\end{subfigure}
\hspace{10mm}
\caption{(a) Intersection scenario with ten agents. (b) Effect of increasing the number of agents on the collision rate.}
\label{fig:scalability}
\end{figure}

To assess the scalability of our methodology, we performed experiments in an intersection scenario, progressively increasing the number of agents within the network. The results, depicted in Figure \ref{fig:scalability} and table \ref{scalability_table}, show the test collision rates (\%) as the agent count varies. As expected, the environment's complexity escalates with each additional agent, making it increasingly difficult to maintain a zero collision rate. Nevertheless, our IntNet approach produces significantly lower collision rates compared to the IC3Net model when the number of agents is increased. The IC3Net model exhibits a growing number of collisions as more agents are added, with rates climbing over 15\% when the environment includes ten agents, highlighting the challenges of scaling in more congested settings, even with communication.

\begin{table}[]
\centering
\caption{Test collision rates (\%) for IntNet and IC3Net across various agent counts.}
\label{scalability_table}
\begin{tabular}{ccc}
\hline
Agent Count & IntNet & IC3Net \\
\hline
2 & \(0.0 \pm 0.12\) & \(0.0 \pm 0.05\) \\
4 & \(1.1 \pm 1.5\) & \(2.2 \pm 1.8\) \\
6 & \(1.7 \pm 1.6\) & \(8.6 \pm 3.5\) \\
8 & \(3.5 \pm 4.5\) & \(12.3 \pm 5.12\) \\
10 & \(5.2 \pm 5.1\) & \(15.4 \pm 7.4\) \\
\hline
\end{tabular}
\end{table}

\subsection{Communication Efficiency}

Our architecture includes a scheduler component that optimises message transmission, decreasing the communication network's load. We evaluated its impact through an experiment in an intersection scenario solely with CAVs, contrasting network performance with and without the scheduler. Without the scheduler, the system defaults to a fully connected network, facilitating communication among all agents within range. Table \ref{table_graph} compares network performance in the intersection environment with the scheduler, revealing that each agent communicates with fewer than half of the agents, on average, while achieving the same success level.

\begin{figure}[h]
\centering
\begin{subfigure}[t]{0.23\textwidth}
    \centering
    \includegraphics[width=\textwidth]{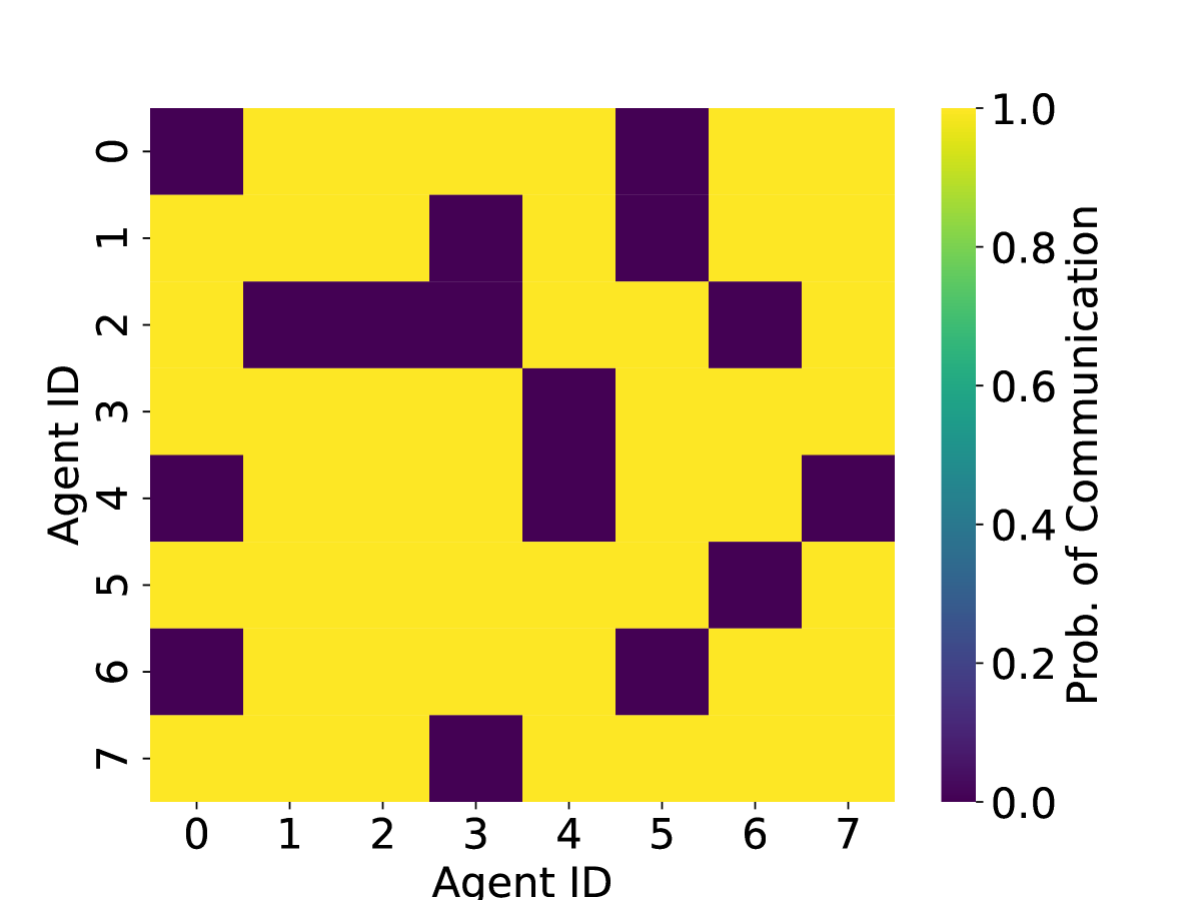}
    \caption{Entering the intersection.}
    \label{fig:comm_entering}
\end{subfigure}
\begin{subfigure}[t]{0.23\textwidth}
    \centering
    \includegraphics[width=\textwidth]{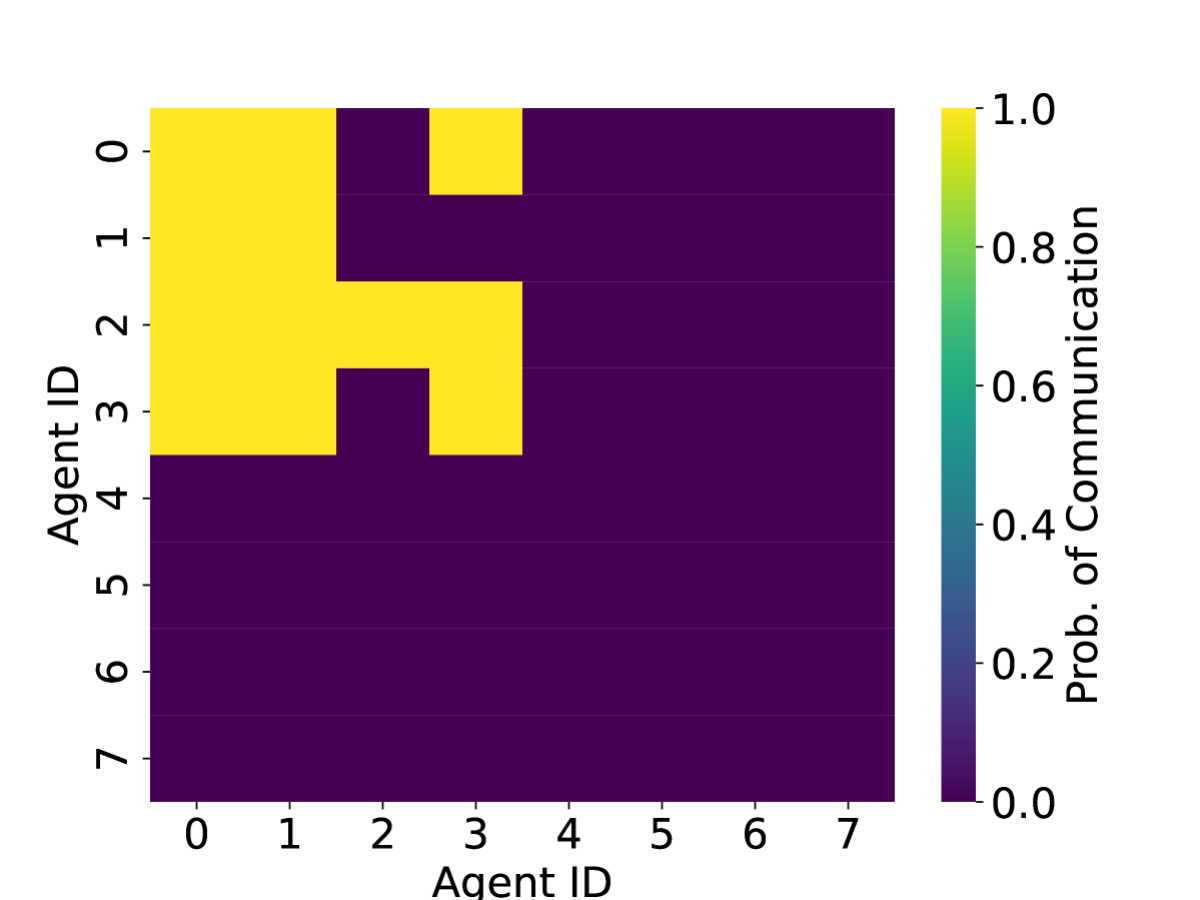}
    \caption{Leaving the intersection}
    \label{fig:comm_leaving}
\end{subfigure}
\hspace{10mm}
\caption{Heatmaps of the communication graphs when entering and leaving the intersection.}
\label{fig:comm_graph}
\end{figure}

Agents communicate only at crucial times for safe navigation, illustrated by Figure \ref{fig:comm_graph}, showing the communication graph in the intersection environment at two key moments. Figure \ref{fig:comm_entering} reveals the graph as agents approach the intersection, and Figure \ref{fig:comm_leaving} displays it as most agents exit, with links in \ref{fig:comm_leaving} (vehicle IDs 0 to 3) indicating vehicles that are still in the intersection.

\begin{table}[h]
\centering
\caption{Effect of the communication scheduler on graph density and success rate.}
\label{table_graph}
\begin{tabular}{ccc}
\hline
 & Graph Density & Success rate (\%) \\
\hline
No Scheduler & \(1.0 \pm 0.0\) & \(86.3 \pm 5.1\) \\
Scheduler  & \(0.41 \pm 0.4\) & \(88.2 \pm 5.8\) \\
\hline
\end{tabular}
\end{table}

\section{CONCLUSIONS}

This study introduces IntNet, a communication-driven MARL framework for cooperative autonomous driving, aimed at enhancing coordination and safety by sharing vehicle intent. We achieve this by incorporating an observation prediction module into a communicative actor-critic architecture. Our model-free training approach uses transitions recorded during episode rollouts to simultaneously train the observation predictor and the policy. The results demonstrate the efficacy of our intent-sharing strategy in fostering safer and more efficient policies in complex urban settings with multiple agents while minimising the communication graph density. Our findings also reveal that sharing intent not only enhances safety and efficiency but also accelerates learning convergence, streamlining the path to optimal policies in complex multi-agent environments. Future work will focus on extending the method's applicability to diverse urban scenarios and varying horizons \(H\), with consideration given to communication delays and packet loss.

\addtolength{\textheight}{-0cm}   



\section*{APPENDIX}

\subsection{Implementation details}
\subsubsection{Environment}

The table below outlines the distinct sets of base parameters used for each of the three scenarios. The reward weights are listed in the order: [$r_{\text{speed}}, r_{\text{col}}, r_{\text{arr}}, r_{\text{merge}}$]. Values represent the weights assigned to each component in the respective scenarios.

\begin{table}[h!]
\centering
\caption{Simulation Parameters for the different scenarios.}
\label{tab:environment_params}
\begin{tabular}{lccc}
\hline
\textbf{Parameter} & \textbf{Highway} & \textbf{Intersection} & \textbf{Lane Merging} \\ \hline
Lanes Count        & 4                & 4                      & 3                      \\
N° Agents          & 4                & 4                      & 3                     \\
N° HDVs            & 4                & 2                      & 2                     \\
Max Steps          & 40               & 15                     & 15                     \\
Speed Range        & [20, 30]         & [0, 9]                 & [20, 30]              \\
Simulation Frequency & 8             & 10                     & 15                     \\
\multirow{2}{*}{Reward Weights} 
                   & [0.4, -5.0,   & [1.0, -5.0,       & [0.4, -1.0,       \\
                   & 0.0, 0.0]             & 1.0, 0.0]                 & 1.0, -0.5]                 \\ \hline
\end{tabular}
\footnotetext{Reward components are listed in the order: $r_{\text{speed}}, r_{\text{coll}}, r_{\text{arr}}, r_{\text{merge}}$. Values represent the weights assigned to each component in the respective scenarios.}
\end{table}

\subsubsection{IntNet details}

The parameters for the proposed method are detailed in the table below:

\begin{table}[h!]
\centering
\caption{IntNet set of hyperparameters.}
\label{tab:hyperparameters}
\begin{tabular}{l c}
\hline
\textbf{Hyper-parameter} & \textbf{Value} \\
\hline
batch size & 200 \\
enc\_hid\_size & 128 \\
gat\_num\_heads & 4 \\
gat\_num\_heads\_out & 1 \\
gat\_hid\_size & 32 \\
gat\_encoder\_out\_size & 32 \\
gamma & 0.9 \\
learning\_rate & 0.0001 \\
value\_coeff & 0.01 \\
\hline
\end{tabular}
\end{table}

\subsubsection{Prediction Horizon $H$}

\begin{figure}[h!]
    \centering
    \includegraphics[width=0.38\textwidth]{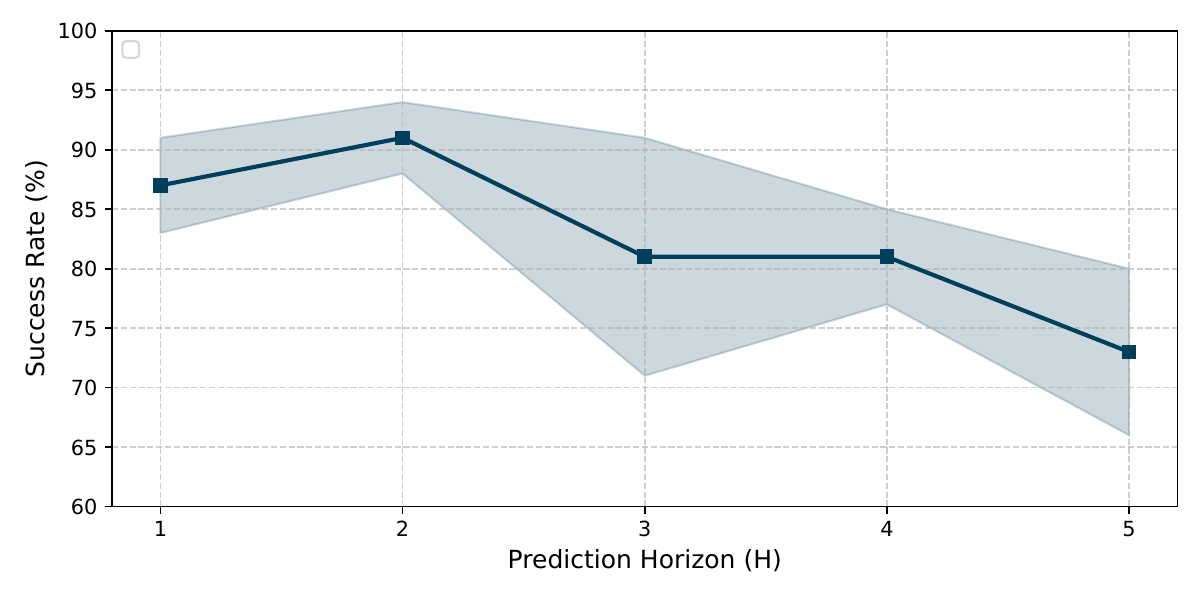} 
    \caption{Average test success rate with varying prediction horizons $H$ in the Lane Merging Scenario.}
    \label{fig:prediction_horizon}
\end{figure}

To determine the optimal prediction horizon $H$, we conducted a preliminary experiment on the Lane Merging Scenario. The architecture was trained with varying $H$ values from 1 to 6, and the average and standard deviation of the test success rate were calculated. The results, shown in Figure \ref{fig:prediction_horizon}, indicate that performance decreased for $H > 2$, with larger variability and diminishing returns. Based on these findings, we selected $H = 2$ for our experiments to balance predictive accuracy and practical utility.

\bibliographystyle{abbrv}
\bibliography{Adv-comm}

\end{document}